\documentclass[runningheads]{llncs}
\usepackage[T1]{fontenc}
\usepackage{listings}
\usepackage{subcaption}
\usepackage{amsmath}
\usepackage{paralist}
\usepackage{graphicx}
\usepackage{multirow}
\usepackage{color}
\usepackage{cite}
\begin{document}

\title{Enhancing Genetic Algorithms with Graph Neural Networks: A Timetabling Case Study}

\titlerunning{Enhancing GAs with GNNs}

\author{Laura-Maria Cornei  \thanks{Corresponding author} \inst{1}\orcidID{0000-0003-1032-0924} \and
Mihaela-Elena Breab\u{a}n\inst{1}\orcidID{0000-0003-4468-3889}}

\authorrunning{Laura-Maria Cornei, Mihaela-Elena Breab\u{a}n}

\institute{Alexandru Ioan Cuza University, Faculty of Computer Science Ia\c{s}i, Romania 
\email{\{laura.cornei,mihaela.breaban\}@info.uaic.ro}}

\maketitle              

\begin{abstract}

This paper investigates the impact of hybridizing a multi-modal Genetic Algorithm with a Graph Neural Network for timetabling optimization. The Graph Neural Network is designed to encapsulate general domain knowledge to improve schedule quality, while the Genetic Algorithm explores different regions of the search space and integrates the deep learning model as an enhancement operator to guide the solution search towards optimality. Initially, both components of the hybrid technique were designed, developed, and optimized independently to solve the tackled task.
Multiple experiments were conducted on Staff Rostering, a well-known timetabling problem, to compare the proposed hybridization with the standalone optimized versions of the Genetic Algorithm and Graph Neural Network. The experimental results demonstrate that the proposed hybridization brings statistically significant improvements in both the time efficiency and solution quality metrics, compared to the standalone methods. To the best of our knowledge, this work proposes the first hybridization of a Genetic Algorithm with a Graph Neural Network for solving timetabling problems.

\end{abstract}

\section{Introduction} \label{introduction}
Genetic Algorithms (GAs), along with other meta-heuristics, have been traditionally applied for solving timetabling problems, as they are able to  provide a set of high-quality and potentially diverse schedules \cite{Ageneticalgorithmforthepersonneltaskreschedulingproblemwithtimepreemption,SolvingamultiobjectiveprofessionaltimetablingproblemusingevolutionaryalgorithmsatMandarineAcademy,Noveloperatorsforquantumevolutionaryalgorithminsolvingtimetablingproblem}. Compared to other meta-heuristics, GAs are well suited for tackling timetabling tasks, as according to Holland's schema theorem \cite{Geneticalgorithmsandadaptation}, the small, high-quality and recurrent schedule patterns can be effectively preserved and combined over generations using crossover.
In the case of real-world timetabling tasks, even small changes in the values of the instances' parameters may lead to completely different solutions. As a result, the process of reapplying GAs, as well as other meta-heuristics, to new instances is usually time consuming.

To avoid this issue, Graph Neural Networks (GNNs) could be used \cite{EverythingisconnectedGraphneuralnetworks}, as they are capable of extracting useful patterns from the training data to quickly 
provide solutions to new test instances, due to their generalization ability.
However, for problems with a huge search space, the optimization process of the GNNs may be prone to getting trapped in local optima \cite{Combinatorialoptimizationandreasoningwithgraphneuralnetworks}.

Consequently, to combine the strengths of both techniques and avoid the aforementioned drawbacks, we propose a hybrid technique in which a pretrained GNN is integrated within a multi-modal GA as a chromosome improvement operator. The pretrained Graph Neural Network would encapsulate general knowledge needed for improving the quality of schedules with possibly different sizes. The Genetic Algorithm would ensure efficient exploration of the search space, performing the concrete steps towards reaching diverse optimal solutions for given test instances. 

The current study brings the following main contributions:
\begin{itemize}
	\item developing and optimizing a Graph Neural Network for solving a well-known timetabling problem, namely Staff Rostering. 
	
	\item designing and optimizing a multi-modal Genetic Algorithm for solving Staff Rostering.
	
	\item integrating the Genetic Algorithm with the Graph Neural Network and evaluating this hybrid approach in comparison to the standalone GNN and GA. The comparative analysis was conducted under multiple settings and was based on several evaluation criteria, including computational time, quality and diversity of the obtained schedules. The experimental results indicate that the proposed hybridization brings statistically significant improvements compared to the standalone optimized versions of the GA and GNN. 
	
\end{itemize}

To the best of our knowledge, our paper is the first to introduce and investigate the impact of integrating a Graph Neural Network with a Genetic Algorithm for solving
timetabling problems. Tests performed on the Staff Rostering problem demonstrate that the developed hybridization leverages the complementary strengths of the GAs and GNNs.

The rest of the paper is organized as follows: section \ref{related_work} discusses related work, while section \ref{nrp_model} introduces the Staff Rostering problem and the used mathematical model. Sections \ref{gnn} and \ref{ga} describe the process of creating and optimizing the Graph Neural Network and the Genetic Algorithm. Section \ref{experiments_results} presents the experimental settings and summarizes the results. Finally, section \ref{conclusions_future_work} proposes future work directions and states the conclusions. 

\section{Related work} \label{related_work}

Over the years, hard timetabling problems have been solved using 
meta-heuristics \cite{Ageneticalgorithmforthepersonneltaskreschedulingproblemwithtimepreemption,Apreventivereactiveapproach,Amultiobjectiveevolutionaryapproachtoprofessionalcoursetimetabling,Noveloperatorsforquantumevolutionaryalgorithminsolvingtimetablingproblem,SolvingamultiobjectiveprofessionaltimetablingproblemusingevolutionaryalgorithmsatMandarineAcademy,Multiobjectivehybridoptimizationsfordesigningcourseschedulesbasedonoperatingcostsandresourceutilization}, mathematical optimization methods \cite{Astochasticintegerprogrammingapproachtoreservestaffschedulingwithpreferences} or a combination of both (math-heuristics) \cite{Nurserosteringwithfatiguemodelling,Ahybridintegerprogrammingandartificialbeecolonyalgorithmforstaffschedulingincallcenters,Amatheuristicbasedsolutionapproachforanextendednurserosteringproblemwithskillsandunits,MaximizingShiftPreferenceforNurseRosteringScheduleUsingIntegerLinearProgrammingandGeneticAlgorithm}.
More recently, GNNs have become increasingly popular for solving Combinatorial Optimization Problems (COPs), including timetabling tasks, due to their generalization ability and special properties (permutation invariance and equivariance)  \cite{Combinatorialoptimizationandreasoningwithgraphneuralnetworks}.

Novel techniques for solving general COPs using Graph Neural Networks 
usually model the problem as a (Mixed) Integer Linear Program ((M)ILP) and then 
encode this information into a bipartite graph associated to the GNN, such that one partition represents the variables of the (M)ILP, and the other the constraints \cite{Agnn-guidedpredict-and-searchframeworkformixed-integerlinearprogramming,RL-MILPSolver:areinforcementlearningapproachforsolvingmixed-integerlinearprogramswithgraphneuralnetworks,Combinatorialoptimizationwithautomatedgraphneuralnetworks}. In order to be able to attain better results, the GNN is usually integrated with other methods, such as Reinforcement Learning (RL) techniques \cite{RL-MILPSolver:areinforcementlearningapproachforsolvingmixed-integerlinearprogramswithgraphneuralnetworks}, mathematical optimization methods \cite{Agnn-guidedpredict-and-searchframeworkformixed-integerlinearprogramming}, trajectory meta-heuristics (Simulated Annealing \cite{Combinatorialoptimizationwithautomatedgraphneuralnetworks})
or population-based meta-heuristics (Large Neighborhood Search \cite{Neurallargeneighborhoodsearch}).
Although applicable to any COP, this kind of approach comes with the disadvantage that increasing the size or complexity of the instances leads to a major increase in the number of vertices and edges in the GNN graph. This further conducts to poorer generalization results, 
as the depth of the GNN cannot be scaled with graph size \cite{Barriersfortheperformanceofgraphneuralnetworksindiscreterandomstructures}
due to potential over-smoothing issues
\cite{oversmoothing}. 
  
Considering the No Free Lunch Theorem \cite{NofreelunchtheoremAreview}, devising GNN-based approaches specific to the employed task represents a promising research line. In case of timetabling tasks, GNNs have been applied in hybridizations with: 
\begin{compactitem}
	\item Reinforcement Learning algorithms, such that the GNN was used as a policy in the RL method. \cite{AutomatedPersonnelSchedulingwithReinforcementLearningandGraphNeuralNetworks,Reinforcementlearningforscalabletraintimetablereschedulingwithgraphrepresentation}
	\item the Large Neighborhood Search meta-heuristic, where the GNN was used as a destroy operator \cite{Alearninglargeneighborhoodsearchforthestaffrerosteringproblem} (idea also presented in \cite{Neurallargeneighborhoodsearch}). 
	\item Mathematical Optimization methods \cite{FasterLargerStrongerOptimallySolvingEmployeeSchedulingProblemswithGraphNeuralNetworks,AcceleratingModelSolvingforIntegratedOptimizationofTimetablingandVehicleSchedulingbasedonGraphConvolutionalNetwork} and classical heuristics \cite{Solvingtherailwaytimetablereschedulingproblemwithgraphneuralnetworks}, which aim to locally improve the solution outputted by the GNN.   
\end{compactitem}

To the best of our knowledge, there are currently no papers integrating a Graph Neural Networks with a Genetic Algorithm for solving timetabling problems. Also, the use of a GNN as an improvement operator within a meta-heuristic is a novel idea. Lastly, there are no current studies evaluating the impact of hybridizing GAs with GNNs for timetabling tasks. 

\section{Staff Rostering - an Integer Programming model} \label{nrp_model}

Staff Rostering \cite{Asurveyofthenurserosteringsolutionmethodologies}, also known as Personnel Timetabling, Staff Scheduling, and Nurse Rostering in the literature, is an NP-hard combinatorial optimization problem involving the assignment of shifts to employees, while satisfying a series of hard and soft constraints. 

Starting from a well-known mathematical model for this problem \cite{Computationalresultsonnewstaffschedulingbenchmarkinstances}, we selected and adjusted several restrictions to fit a real-world scheduling scenario and ensure compliance with our country's work legislation. Let $E$ be the number of employees, $D$ the number of days in the planning horizon, and
$S$ the set of shifts, such that there are three 8-hour shifts in total: morning, afternoon and night. Let $x_{eds}$ be an indicator variable, taking a value of $1$ if employee $e$ is working shift $s$ on day $d$, and $0$ otherwise. 

The final model includes the following hard constraints:
\begin{compactitem}
	\item C1. Each employee works at most one shift per day: 
	\begin{equation}
		\begin{aligned}
			\sum_{s \in S}{x_{eds}} \leq 1, \forall e \in \{1,...,E\}, d \in \{1,...,D\}
		\end{aligned}
	\end{equation}
	\item C2. Morning shifts cannot be preceded by night shifts:
	\begin{equation}
		\begin{aligned}
		x_{eds} + x_{e(d+1)r} \leq 1, \forall e \in \{1,...,E\},\\ d \in \{1,..., D-1\}, s = night, r = morning
	   \end{aligned}
	\end{equation}
	\item C3. Each employee has to work at least $b^{min}$ hours and at most $b^{max}$ hours:
	\begin{equation}
		\begin{aligned}
			b^{min} \leq \sum_{d \in \{1,...,D\}}\sum_{s \in S} 8 \cdot x_{eds}  \leq b^{max}, \forall e \in \{1,...,E\} 
		\end{aligned}
	\end{equation}
	\item C4. Each employee has to work at most $c^{max}$ consecutive shifts:
	\begin{equation}
		\begin{aligned}
			\sum_{d = t}^{t + c^{max}}\sum_{s \in S} x_{eds}  \leq c^{max}, \forall e \in \{1,...,E\}, t \in \{1, ..., D - c^{max}\}
		\end{aligned}
	\end{equation}
	\item C5. When taking days off, each employee has to rest for at least $o^{min}$ days:  
	\begin{equation}
		\begin{aligned}
			(1 - \sum_{s \in S} x_{eds}) + \sum_{j = d + 1}^{d + t}\sum_{s \in S} x_{ejs} + (1 - \sum_{s \in S} x_{e(d+t+1)s}) > 0, \\ \forall e \in \{1,...,E\}, t \in \{1,...,o^{min}-1\}, d \in \{1,..., D-(t+1)\} 
		\end{aligned}
	\end{equation}
\end{compactitem}

The objective function minimizes the sum of penalties associated to the understaffing ($v^{min}$) and overstaffing ($v^{max}$) situations respectively, as well as the penalties corresponding to the work preferences of the employees ($p_{ed}$). Specifically, $p_{ed}$ is an indicator variable taking a value of 1 when employee $e$ prefers not to work on day $d$, and $0$ otherwise. $P$ represents the total number of penalties of type $p_{ed}$.
\begin{equation*} \label{eqobjfunc}
	\begin{aligned}
		{} &
		\min \left( 
		{\sum\limits_{d \in \{1,...,D\}}\sum\limits_{s \in S} {y_{ds}v^{min}}} + 
		{\sum\limits_{d \in \{1,...,D\}}\sum\limits_{s \in S} {z_{ds}v^{max}}} 		
		+
		{ \sum\limits_{e \in \{1,...,E\}} \sum\limits_{d \in \{1,...,D\}}\sum\limits_{s \in S} {p_{ed} x_{eds}}}
		\right) 
	\end{aligned}
\end{equation*} where:
\begin{equation}
	\begin{aligned}
		y_{ds} = {\max (0, u_{ds} -\sum\limits_{e \in \{1,...,E\}} x_{eds})},\quad z_{ds} = {\max (0, \sum\limits_{e \in \{1,...,E\}} x_{eds} - u_{ds})} 
	\end{aligned}
\end{equation}

\section{Graph Neural Network} \label{gnn}

\subsection{GNN learning procedure} \label{gnn_learning}
\begin{figure}
	\includegraphics[width=0.9\textwidth]{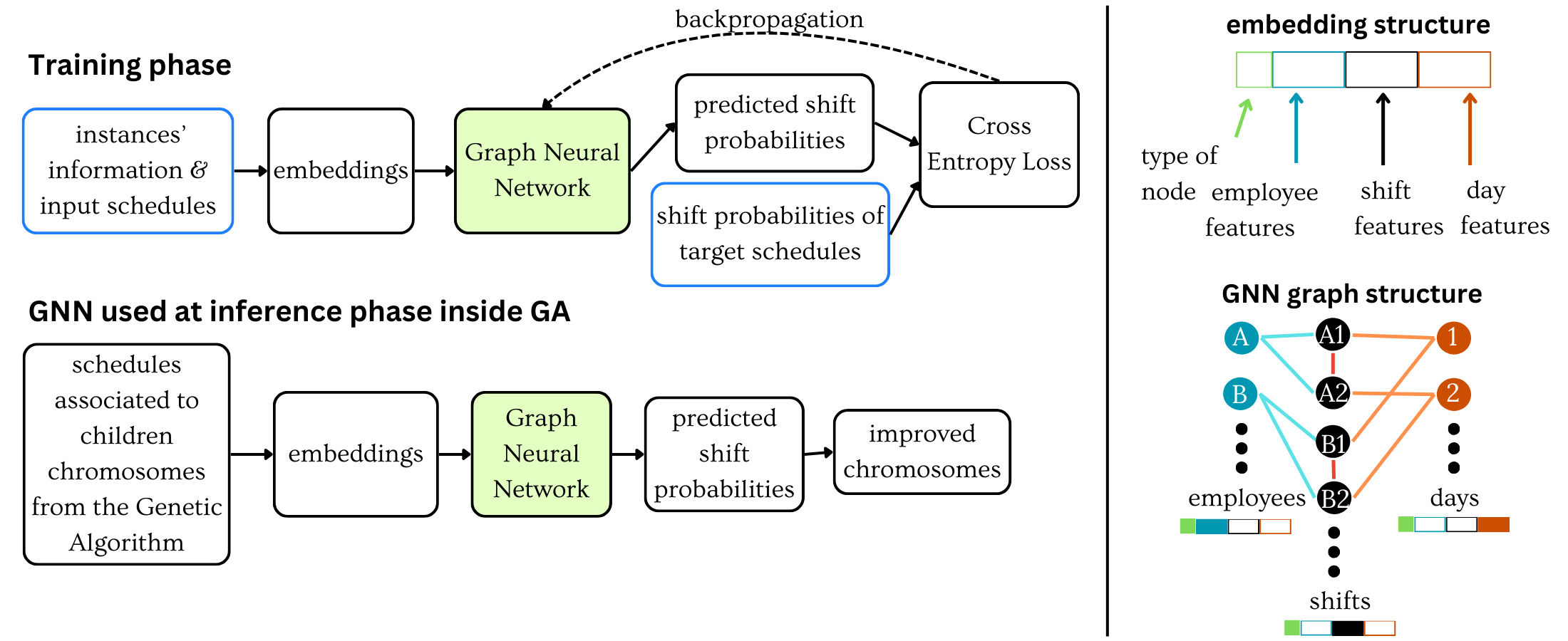}
	\caption{Overview of the GNN learning procedure (left side) and structure of the graph and the embeddings of the used GNN (right side)} \label{fig1}
\end{figure}

Figure \ref{fig1} illustrates, on the left side, an overview of the GNN learning procedure and, on the right side, the structure of the graph and the embeddings associated to the employed GNN. 

The used Graph Neural Network acts as a solution improvement operator, such that given as input a batch of schedules, it predicts timetables with a considerably higher quality with respect to the objective function. Although the GNN was trained on pairs of (input, target) schedules corresponding to multiple instances, each pair of timetables was associated to the same instance. As loss function, the Cross Entropy computed between the probability distributions of the predicted and target shifts was used.

To avoid overfitting, we employed an early stopping procedure such that the training was halted when a number of $patience$ epochs elapsed without a significant improvement in the validation loss.
The validation loss measured the quality of the generated schedules by minimizing the difference between the number of unfeasible and optimal predicted schedules, divided by the total number of timetables. In case this value remained constant, the loss aimed to minimize the Mean Squared Error between the objective function values of the feasible suboptimal and optimal schedules.

At test time, the use of the GNN within the Genetic Algorithm involved extracting the GNN embeddings from the schedules associated to each chromosome, feeding them (only once) to the network, and then formatting the output to obtain the resulting chromosomes.

\subsection{Datasets generation} \label{gnn_datasets}

To train the GNN we needed to collect pairs of schedules $(s,s')$, such that $s'$ had a better (lower) objective function value than $s$. To construct the datasets, we first generated $250$ problem instances parameterized by $100$ persons and $7$ days, and computed an optimal solution for each of them  using the Gurobi solver \cite{gurobi}. For all instances, the following problem parameters were set to the indicated values: $b^{min}=32,b^{max}=48,c^{max}=5,o^{min}=2,v^{min}=100,v^{max}=1$. The preferred coverage for each day $d$ and shifts $s$ ($u_{ds}$) was set to the floor division between the number of employees $E$ and the number of shifts $\lvert S \rvert$. The shift off request penalties for each employee $e$ and day $d$ ($p_{ed}$) randomly took values from the set $\{0,1\}$. Next, we altered via random shift changes the optimal solutions to obtain for each of them $250$ unique feasible schedules, resulting in a total of $62500$ feasible timetables. Then, we applied random shift mutations to the feasible schedules to obtain one unfeasible timetable for each feasible one. 

The final dataset included pairs of unfeasible-feasible and feasible-optimal schedules. We additionally created other types of datasets, containing for example pairs of unfeasible-optimal and feasible-optimal timetables; however, the initially proposed version yielded the best results, so we adopted it. 
On this dataset, we applied an 80/10/10 split for training, validation and testing.

\subsection{GNN structure, model and embeddings} \label{gnn_embeddings}

As illustrated on the right side of Figure \ref{fig1}, the GNN graph contains three types of vertices corresponding to employees, shifts and days. A shift node  was created for every (employee, day) pair and was connected to the related employee and day vertices via bidirectional edges. Shift nodes linked to the same employee vertex but to consecutive day vertices were 
connected as well, to encourage the flow of temporal information.

Using the PyTorch Geometric library \cite{PyTorchGeometric}, we created a Graph Neural Network architecture consisting of $nb\_layers\_conv$ TransformerConv GNN layers \cite{MaskedlabelpredictionUnifiedmessagepassingmodelforsemisupervisedclassification}, followed by a multilayer perceptron (MLP) with $nb\_layers\_MLP$ layers. To generate an output starting from the input embeddings, the network first applied the GNN convolutional layers to produce new node representations for the shift, employee and day nodes. The shift representations were further concatenated with the corresponding employee and day ones and the resulting embeddings were passed through the MLP to obtain for each (employee $e$, day $d$) pair the probabilities of $e$ being assigned each type of shift (including the rest one) on day $d$. Given this generated output, the reconstruction of the resulting schedule was straightforward, by selecting, for each (employee $e$, day $d$) pair, the shift value with the highest probability. The used GNN was heterogeneous, such that different types of relations (employee-shift, shift-day, shift-shift) had associated separate Message Passing functions \cite{Modelingrelationaldatawithgraphconvolutionalnetworks}.   

The features associated with each node in the GNN graph contained, in the case of the hard constraints, flags signaling their infringement and, in the case of both hard and soft restrictions, normalized values signifying the degree to which they were or were not satisfied. The final embeddings were formed, as highlighted in Figure \ref{fig1}, by concatenating 2 digits representing the type of the GNN node with three kinds of features (associated with employees, shifts, or days), such that zero values were placed in case the feature type did not match the the node type. The employee features included information regarding the $C3$ constraint, the day features retained the normalized understaffing and overstaffing soft penalties, while the shift features integrated the one hot encoding of the shifts, as well as information concerning all the other remaining restrictions.

\subsection{GNN hyper-parameter tuning} \label{hyper-param_tuning}

To optimize the designed Graph Neural Network, we performed a random search hyper-parameter tuning by varying the following hyper-parameters:
\begin{compactitem}
	\item $batch\_size \in \{64, 128, 256\}$ (training batch size)
	\item $lr \in \{0.01, 0.001\}$ (learning rate)
	\item $opt\_weight\_decay \in \{0.001, 0.0001\}$ (optimizer weight decay)
	\item $dropout\_conv \in \{0, 0.1, 0.2\}$ (dropout for GNN convolutional layers)
	\item $dropout\_MLP \in \{0, 0.1, 0.2\}$ (dropout for MLP layers)
	\item $nb\_layers\_conv \in \{3,4\}$ (number of GNN convolutional layers)
	\item $nb\_layers\_MLP \in \{3,4,5\}$ (number of MLP layers)
	\item $nb\_heads \in \{4, 8\}$ (number of GNN attention heads)
	\item $patience \in \{20,30,40\}$ (early stopping patience)
\end{compactitem}

We used the AdamW optimizer due to its generalization ability \cite{TowardsunderstandingconvergenceandgeneralizationofAdamW}.
The best obtained configuration out of 40 trials was: $batch\_size=64,  opt\_weight\_decay= 0.0001,lr=0.001, dropout\_conv=0,dropout\_MLP=0,nb\_layers\_conv=4,nb\_layers\_MLP=4,nb\_heads = 8,patience=30$.
To select the best configuration, we trained the GNN 30 times for each of them and compared the averaged results on the test dataset.

\section{Genetic Algorithm} \label{ga}

\subsection{Overview of the algorithm} \label{overview_ga}

The proposed Genetic Algorithm was used to find optimal and near-optimal diverse solutions for 
a given problem instance. To provide statistically significant results, we ran the algorithm $10$ times on each of the $50$ randomly generated instances. These instances had the same size (100 persons and 7 days) and were generated following the same procedure as the one employed for training the GNN, presented in section \ref{gnn_datasets}. However, we ensured that the generated instances were different from those used in the GNN learning and testing process. 

The pseudocode for the employed Genetic Algorithm, presented in Listing \ref{pseudocode}, integrates the classical GA operators (selection, crossover, mutation) and, optionally, the GNN-based improvement operator. Next, we discuss each component of the Genetic Algorithm.

\begin{lstlisting}[caption={Genetic Algorithm Pseudocode},columns=fullflexible,language=Python,label=pseudocode,belowskip=-3pt,aboveskip=1pt]
def genetic_algorithm(pop_size, stop_cond_version, nb_max_epochs, 
                       max_patience, probab_crossover, probab_mutation,
                       min_prob_greedy, use_GNN):         
pop = get_init_population(pop_size)
epoch = 0 
patience = 0
prob_greedy = min_prob_greedy
fitness_and_pen_info = get_fitness_and_pen_info(pop)

while not stop_alg(stop_cond_version, epoch, nb_max_epochs, 
                    patience, max_patience, fitness_and_pen_info):
	offspring = crossover_all(pop, probab_crossover)
	offspring = mutation_all(offspring, probab_mutation)
	if use_GNN:
		offspring = GNN_improvement(offspring)
	distances = calc_crowding_distances(pop, offspring)
	matches = find_matchings(distances)
	prob_greedy = update_prob_greedy(prob_greedy, min_prob_greedy, epoch)
	pop = selection(pop, offspring, matches, prob_greedy)
	fitness_and_pen_info = get_fitness_and_pen_info(pop)
	patience = update_patience(patience, fitness_and_pen_info)
	epoch += 1
\end{lstlisting}

\subsubsection{Solution representation, initial population and fitness function}

A chromosome in the population encodes a schedule as a matrix, where each row corresponds to a person, each column to a day, and each entry maps the shift assigned to a person in a day. 
To obtain the initial population, we randomly generated $pop\_size$ chromosomes using the $get\_init\_population$ function, allowing both feasible and unfeasible schedules to be part of the initial population. 

The fitness value of a chromosome minimizes the violation of the hard and soft constraints, being computed as:
\begin{equation}
	fitness\_value = MAX\_FITNESS\_VALUE - (pen\_hard + pen\_soft)
\end{equation}

The $pen\_hard$ and $pen\_soft$ scores represent the sums of normalized soft and hard penalties associated with the current schedule; each violation of a hard constraint resulted in a penalty of $1$, while the violation of a soft constraint produced a penalty value in the range $(0,1)$. In $fitness\_and\_pen\_info$, we retained information concerning the fitness values, the hard penalties, and the normalized and unnormalized soft penalties associated with the current population. $MAX\_FITNESS\_VALUE$ represents an upper bound estimation for the maximum fitness value and was computed as $
nb\_persons * nb\_days * (nb\_shifts + 1) + nb\_shifts * nb\_days + nb\_persons$. Consequently, an optimal solution has an associated fitness value of $2921$.

\subsubsection{Stopping conditions} \label{stop_cond_ga}

In the experiments section \ref{experiments_results}, we used two versions of the GA, differing based on the employed stopping conditions. In the first version of the Genetic Algorithm denoted as $GA\_v1$ (having $stop\_cond\_version$ set to $1$), the algorithm was halted either upon completion of $nb\_max\_epochs$ or upon reaching an optimal solution. We considered a schedule to be optimal if it had a hard penalty equal to $0$ and a soft (unnormalized) penalty equal to the minimum soft penalty associated with the current instance. In the second version of the Genetic Algorithm, denoted as $GA\_v2$ (having $stop\_cond\_version$ set to $2$), the algorithm was stopped either after exceeding a maximum number of epochs $nb\_max\_epochs$, or after exceeding a maximum patience $max\_patience$, indicating the number of epochs passed with no improvement in the best fitness value. The information regarding the best fitness value per epoch was retained in $fitness\_and\_pen\_info$, and the $update\_patience$ function updated the current patience value. 

\subsubsection{Variation operators}

The $crossover\_all$ function creates at first $pop\_size \div 2$ pairs of parents and then applies the crossover operator with a probability of $probab\_crossover$ on each pair to obtain two offspring chromosomes. We chose $pop\_size$ to be an even number to avoid the extra step of pairing up the last chromosome. We implemented and tested multiple ways of performing crossover between two parents $p1$ and $p2$: 
\begin{compactitem}
	
	\item $cx\_one\_line$ randomly chooses one line from $p1$ and one line from $p2$ and swaps them.
	\item $cx\_one\_line\_partially$ randomly chooses a piece of line from $p1$ and a piece of line from $p2$ and swaps them. Both pieces have the same length, but possibly different starting points in the lines. 
 
\end{compactitem}

Each chromosome $ch$ obtained via crossover is further mutated with a probability of $probab\_mutation$ in the $mutation\_all$ function. 
We also tested multiple mutation operators, namely:
\begin{compactitem}
	\item $mut\_swap\_shifts$ and $mut\_change\_shift$. The first swapped two random shifts from $ch$, while the second randomly changed the value of an aleatory chosen shift. 
	\item $mut\_change\_shift\_penalty$, which randomly modified the shift $s$ whose value $v$ induced the highest penalty. This penalty was computed as the sum of hard and normalized soft penalties associated to the hard and soft constraints violated when assigning $v$ to $s$. 
\end{compactitem}

If the $use\_GNN$ flag is set to $True$, the GNN-based operator is applied  to the offspring obtained via mutation, as described in section \ref{gnn_learning}.  

\subsubsection{Crowding distances computation and selection}

To enhance and preserve diversity in the population, we employed a niching mechanism based on probabilistic crowding \cite{Thecrowdingapproachtonichingingeneticalgorithms}. The $calc\_crowding\_distances$ function is used to compute the crowding distances between the initial parent chromosomes and the offspring generated after applying mutation and crossover. The crowding distance between two chromosomes is defined as the number of positions at which the corresponding shift values differ in the associated schedules. Next, the $find\_matchings$ function, taking as input these distances, performs a one-to-one matching between parents and offspring based on the linear sum assignment algorithm \cite{scipy}. Therefore, the offspring chromosomes tended to be assigned to resembling parents, and vice versa. 

Finally, the tournament selection function selects from each (parent $p$, offspring $c$) pair the winning chromosome as follows: with a probability equal to $prob\_greedy$, the winning chromosome is deterministically chosen to be the one with the highest fitness value;  otherwise, a probabilistic rule is used such that  the offspring is selected as the winner with a probability equal to $fitness(c)/(fitness(c)+fitness(p))$ \cite{Thecrowdingapproachtonichingingeneticalgorithms}. The parameter $prob\_greedy$ was initially set to $min\_prob\_greedy$ and its value was linearly increased up to $1.0$, in the $update\_prob\_greedy$ function, as generations evolve. 

\subsection{Optimizing the Genetic Algorithm}
The presented Genetic Algorithm was implemented in Python in order to integrate the GNN-based operator. To speed up the GA functions, we used, whenever possible, the Numba just-in-time compiler \cite{numba}. 
We experimentally tested multiple combinations of parameters and operators for the Genetic Algorithm. The final configuration that provided the best results, out of all trials, was the following: $pop\_size=200$,  $nb\_max\_epochs=50000$, $max\_patience=3000$, $probab\_crossover=0.5$, $probab\_mutation=1.0$, $min\_prob\_greedy=0.4$. The crossover function was randomly chosen with equal probability between $cx\_one\_line$ and $cx\_one\_line\_partially$. The mutation operator applied $mut\_swap\_shifts$, $mut\_change\_shift$ and $mut\_change\_shift\_penalty$ with probabilities 0.2, 0.2 and 0.6 respectively. Although fixing the mutation probability to $1.0$ may appear unconventional, the perturbations induced by the mutation operators are very small ($2$ shifts are changed at most out of $700$) and the penalty-based mutation randomly changes a shift that is very likely to be modified in a chromosome having a high fitness.

\section{Experiments and results} \label{experiments_results}

We evaluated the Graph Neural Network (GNN), the Genetic Algorithm (GA) and their hybridization (GA+GNN) on the same test dataset formed out of $50$ randomly chosen instances. For the GA and GA+GNN versions, we performed $10$ runs for each instance, as explained in section \ref{overview_ga}. All experiments were executed on a Desktop with the following specifications: Intel core ultra 9 285k, NVIDIA GeForce RTX 5080 16GB, 192GB DDR5, 4TB SSD.

We first evaluated the performance of the GNN on the test data set 
and we obtained the following mean values (the associated standard deviations are reported in parentheses): $0.2665$ $(\pm0.0169)$ for the percent of optimal generated schedules,  $0.6486$ $(\pm0.0519)$ for the percent of feasible suboptimal obtained timetables, and $0.0849$ $(\pm0.0365)$ for the percent of unfeasible generated schedules.
Therefore, although obtaining a large ratio of feasible suboptimal timetables, the GNN wasn't able to generate an optimal solution for many of the test instances. In contrast, as we will further see, the GA and GA+GNN versions were able to find optimal solutions for all instances. 

Next, we tested the performance of the GA and GA+GNN methods. According to section \ref{stop_cond_ga}, to provide better insights, we tested two versions of the Genetic Algorithm differing based on the stopping condition: $GA\_v1$ and $GA\_v2$. Method $GA\_v1$ was halted when encountering an optimal solution, while method $GA\_v2$ was terminated based on a best fitness patience, fixed to a value ($3000$) high enough to encourage the convergence of the algorithm.

\begin{figure}[!ht]
	
	\begin{subfigure}[h]{0.45\linewidth}
		\includegraphics[width=1.1\linewidth]{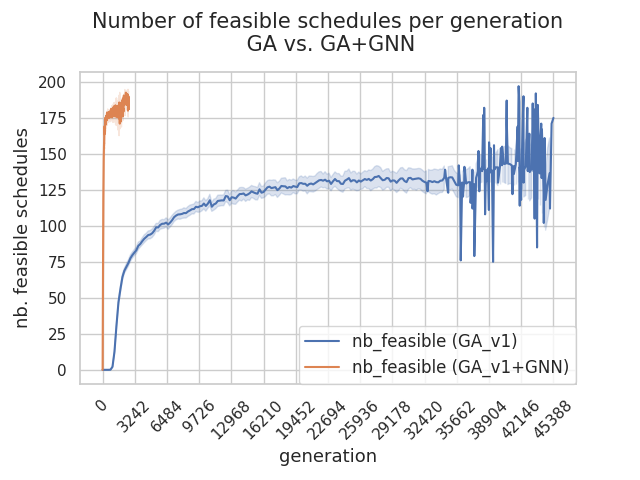}
	\end{subfigure}
	\hfill
	\begin{subfigure}[h]{0.45\linewidth}
		\includegraphics[width=1.1\linewidth]{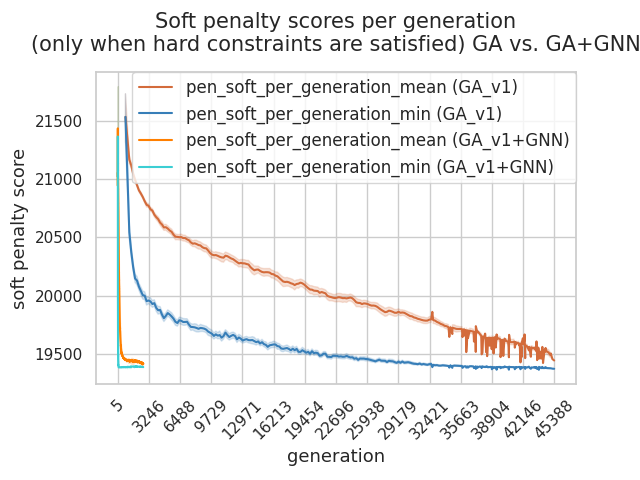}
	\end{subfigure}
	
	\begin{subfigure}[h]{0.45\linewidth}
		\includegraphics[width=1.1\linewidth]{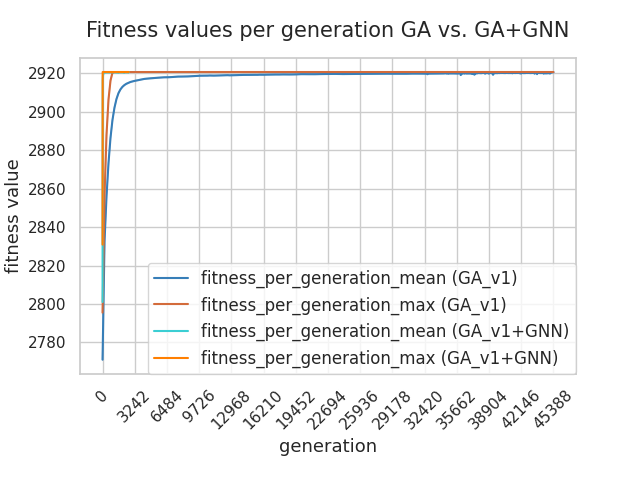}
		
	\end{subfigure}
	\hfill
	\begin{subfigure}[h]{0.45\linewidth}
		\includegraphics[width=1.1\linewidth]{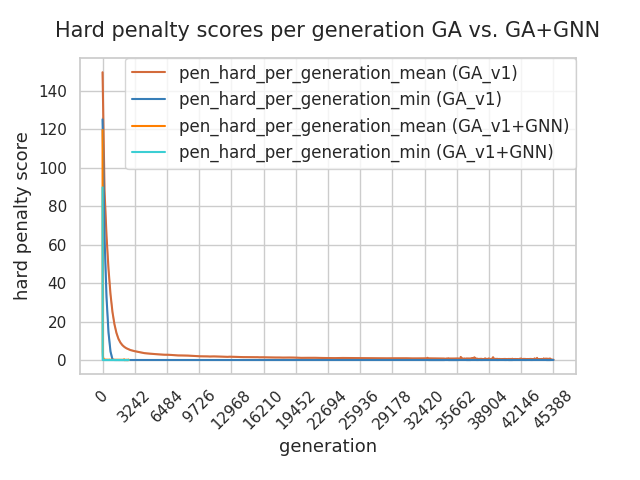}
		
	\end{subfigure}
	
	\begin{subfigure}[h]{0.45\linewidth}
		\includegraphics[width=1.1\linewidth]{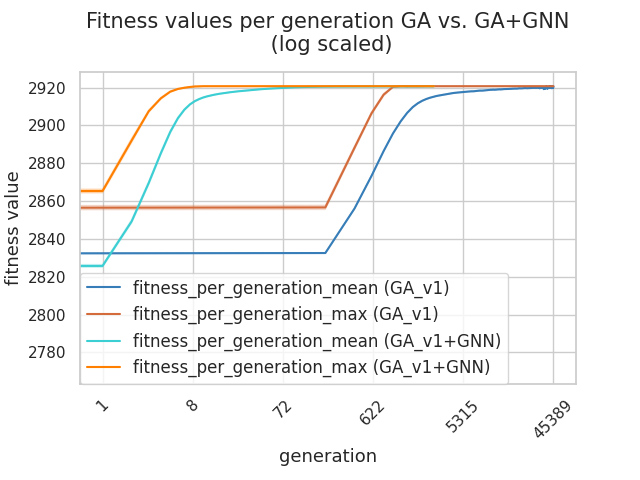}
	\end{subfigure}
	\hfill
	\begin{subfigure}[h]{0.45\linewidth}
		\includegraphics[width=1.1\linewidth]{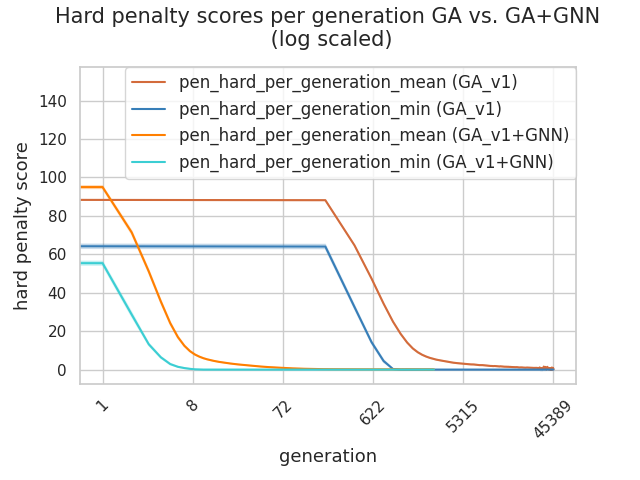}
	\end{subfigure}

	\caption{Comparative evaluation of the $GA\_v1$ and $GA\_v1+GNN$ methods for the  fitness, hard \& soft penalty scores, and number of feasible schedules metrics}
	\label{expplotsv1}
\end{figure}

Figure \ref{expplotsv1} includes several plots highlighting the comparative evaluation of the $GA\_v1$ and $GA\_v1+GNN$ techniques with respect to: the mean and maximum fitness values, the mean and minimum soft \& hard penalties, and the average number of feasible schedules. The plots illustrate the mean per generation for each metric, as well as the confidence interval at a significance level of $0.05$. In some cases, this interval was extremely small and thus may not be clearly visible. When computing the statistics for the soft penalties, we only considered those scores associated with feasible schedules, to provide a more reliable indicator. For the fitness and hard penalty metrics we additionally provide log scaled plots for better visualization.  

As shown in Figure \ref{expplotsv1}, the $GA\_v1+GNN$ version reaches an optimal solution significantly faster (with respect to the number of generations) than $GA\_v1$. As the search progressed towards an optimal solution, the variability in the number of feasible schedules increased. This behavior may be explained by the increased difficulty in maintaining feasibility in this region, combined with the need to explore infeasible schedules in order to reach the optimal solution. 
Although not included, plots comparing $GA\_v2$ and $GA\_v2+GNN$ provided similar insights.

Table \ref{quality_tab} reports the mean values and standard deviations (reported in parenthesis) of multiple metrics, obtained when applying all four methods:  $GA\_v1$, $GA\_v1+GNN$, $GA\_v2$, $GA\_v2+GNN$. Besides the previously mentioned metrics, we included the average number of optimal solutions, the mean and maximum crowding distance, the average total time in minutes per run and the mean stopping generation number. A high value of the crowding distance indicates a diverse population, which is desirable. 
The means and standard deviations were computed by aggregating each metric's values at the generation corresponding to the best fitness value within the last $1000$ generations.

To compare $GA\_v1$ and $GA\_v1+GNN$, as well as $GA\_v2$ and $GA\_v2+GNN$, we applied the Welch’s t-test to each metric's results and highlighted in bold the winning outcomes, whenever the difference was statistically significant at the $95\%$ confidence level.

\begin{table}[!ht]
	\caption{Comparative evaluation of the $GA\_v1$, $GA\_v1+GNN$, $GA\_v2$ and $GA\_v2+GNN$ methods for multiple metrics} 
	\label{quality_tab}
	\begin{tabular}{|ll|l|l|l|l|}
		\hline
		\multicolumn{2}{|l|}{\begin{tabular}[c]{@{}l@{}}Metric /\ Method\end{tabular}} &
		GA\_v1 &
		GA\_v1 + GNN &
		GA\_v2 &
		GA\_v2 + GNN \\ \hline
		\multicolumn{1}{|l|}{\multirow{2}{*}{Fitness value}} &
		mean &
		\textbf{\begin{tabular}[c]{@{}l@{}}2920.2208\\ (0.3655)\end{tabular}} &
		\begin{tabular}[c]{@{}l@{}}2919.7809\\ (1.4256)\end{tabular} &
		\begin{tabular}[c]{@{}l@{}}2920.5296\\ (0.1464)\end{tabular} &
		\textbf{\begin{tabular}[c]{@{}l@{}}2920.7144\\ (0.0047)\end{tabular}} \\ \cline{2-6} 
		\multicolumn{1}{|l|}{} &
		max &
		\begin{tabular}[c]{@{}l@{}}2920.7177\\ (0.0006)\end{tabular} &
		\begin{tabular}[c]{@{}l@{}}2920.7178\\ (0.0006)\end{tabular} &
		\begin{tabular}[c]{@{}l@{}}2920.7165\\ (0.0017)\end{tabular} &
		\textbf{\begin{tabular}[c]{@{}l@{}}2920.7187\\ (0.0007)\end{tabular}} \\ \hline
		\multicolumn{1}{|l|}{\multirow{2}{*}{Soft penalty}} &
		min &
		\begin{tabular}[c]{@{}l@{}}19380.6833\\ (7.7819)\end{tabular} &
		\begin{tabular}[c]{@{}l@{}}19380.4833\\ (7.7841)\end{tabular} &
		\begin{tabular}[c]{@{}l@{}}19388.8333\\ (13.7633)\end{tabular} &
		\textbf{\begin{tabular}[c]{@{}l@{}}19373.65\\ (7.9550)\end{tabular}} \\ \cline{2-6} 
		\multicolumn{1}{|l|}{} &
		mean &
		\textbf{\begin{tabular}[c]{@{}l@{}}19562.3814\\ (84.3713)\end{tabular}} &
		\begin{tabular}[c]{@{}l@{}}20045.8460\\ (513.1283)\end{tabular} &
		\begin{tabular}[c]{@{}l@{}}19550.3685\\ (120.4645)\end{tabular} &
		\textbf{\begin{tabular}[c]{@{}l@{}}19380.5479\\ (7.9052)\end{tabular}} \\ \hline
		\multicolumn{1}{|l|}{\multirow{2}{*}{Hard penalty}} &
		min &
		\begin{tabular}[c]{@{}l@{}}0.0 \\ (0.0)\end{tabular} &
		\begin{tabular}[c]{@{}l@{}}0.0\\ (0.0)\end{tabular} &
		\begin{tabular}[c]{@{}l@{}}0.0\\ (0.0)\end{tabular} &
		\begin{tabular}[c]{@{}l@{}}0.0\\ (0.0)\end{tabular} \\ \cline{2-6} 
		\multicolumn{1}{|l|}{} &
		mean &
		\textbf{\begin{tabular}[c]{@{}l@{}}0.493\\ (0.3649)\end{tabular}} &
		\begin{tabular}[c]{@{}l@{}}0.9263\\ (1.4214)\end{tabular} &
		\begin{tabular}[c]{@{}l@{}}0.1834\\ (0.1447)\end{tabular} &
		\textbf{\begin{tabular}[c]{@{}l@{}}0.0034\\ (0.0047)\end{tabular}} \\ \hline
		\multicolumn{1}{|l|}{\begin{tabular}[c]{@{}l@{}}Number of\\ feasible \\ schedules\end{tabular}} &
		mean &
		\textbf{\begin{tabular}[c]{@{}l@{}}145.5333\\ (31.5709)\end{tabular}} &
		\begin{tabular}[c]{@{}l@{}}132.8\\ (54.1553)\end{tabular} &
		\begin{tabular}[c]{@{}l@{}}173.8857\\ (16.2386)\end{tabular} &
		\textbf{\begin{tabular}[c]{@{}l@{}}199.8\\ (0.4472)\end{tabular}} \\ \hline
		\multicolumn{1}{|l|}{Nb. optimal schedules} &
		mean &
		\begin{tabular}[c]{@{}l@{}}1.0\\ (0.0)\end{tabular} &
		\begin{tabular}[c]{@{}l@{}}1.0 \\ (0.0)\end{tabular} &
		\begin{tabular}[c]{@{}l@{}}20.5428\\ (35.7008)\end{tabular} &
		\textbf{\begin{tabular}[c]{@{}l@{}}156.8292\\ (25.7235)\end{tabular}} \\ \hline
		\multicolumn{1}{|l|}{\multirow{2}{*}{Crowding distance}} &
		mean &
		\textbf{\begin{tabular}[c]{@{}l@{}}471.8697\\ (8.9183)\end{tabular}} &
		\begin{tabular}[c]{@{}l@{}}453.3819\\ (9.4435)\end{tabular} &
		\textbf{\begin{tabular}[c]{@{}l@{}}471.7008\\ (17.4030)\end{tabular}} &
		\begin{tabular}[c]{@{}l@{}}439.6751\\ (6.0230)\end{tabular} \\ \cline{2-6} 
		\multicolumn{1}{|l|}{} &
		max &
		\textbf{\begin{tabular}[c]{@{}l@{}}539.0\\ (8.7662)\end{tabular}} &
		\begin{tabular}[c]{@{}l@{}}514.2666\\ (10.0150)\end{tabular} &
		\textbf{\begin{tabular}[c]{@{}l@{}}542.8833\\ (22.9539)\end{tabular}} &
		\begin{tabular}[c]{@{}l@{}}498.9166\\ (7.4339)\end{tabular} \\ \hline
		\multicolumn{1}{|l|}{Total time 1 run (mins.)} &
		mean &
		\begin{tabular}[c]{@{}l@{}}23.1038\\ (11.0696)\end{tabular} &
		\textbf{\begin{tabular}[c]{@{}l@{}}2.2620\\ (3.2678)\end{tabular}} &
		\textbf{\begin{tabular}[c]{@{}l@{}}17.9557\\ (6.1041)\end{tabular}} &
		\begin{tabular}[c]{@{}l@{}}69.9784\\ (20.7085)\end{tabular} \\ \hline
		\multicolumn{1}{|l|}{Stop generation} &
		mean &
		\begin{tabular}[c]{@{}l@{}}40479.75\\ (2987.77)\end{tabular} &
		\textbf{\begin{tabular}[c]{@{}l@{}}400.15\\ (579.39)\end{tabular}} &
		\begin{tabular}[c]{@{}l@{}}34032.25\\ (11635.94)\end{tabular} &
		\textbf{\begin{tabular}[c]{@{}l@{}}12956.58\\ (3852.54)\end{tabular}} \\ \hline
	\end{tabular}
\end{table}

 On average, a run with $GA\_v1+GNN$ was more than $10$ times faster than a run with $GA\_v1$ and this difference in terms of efficiency was also reflected in the average value of the stop generation ($\sim400$ vs. $\sim40480$). 
 For the other metrics, the average values of $GA\_v1+GNN$  were usually lower or slightly lower than those of  $GA\_v1$.
 However, the $GA\_v1+GNN$ version obtained results that were statistically indistinguishable from those of $GA\_v1$ for three metrics indicating the quality of the best solutions: maximum fitness value, minimum soft penalty, minimum hard penalty. 
 In conclusion,  $GA\_v1+GNN$ represents a significantly faster alternative to  $GA\_v1$, such that the quality of best schedules obtained by both methods was statistically comparable.

 The $GA\_v2+GNN$ obtained considerably better results compared to $GA\_v2$ for all metrics except for the minimum hard penalty, for which the results were comparable, and for the crowding distance, for which the results were worse. A notable aspect is the increase brought by $GA\_v2+GNN$ in the average number of optimal schedules when compared to the classical GA ($\sim156.8$ vs $\sim20.5$). Also, almost all schedules associated to the chromosomes from the $GA\_v2+GNN$'s population were feasible (on average $199.8$ out of $200$). 
 These improvements brought by $GA\_v2+GNN$ came with a trade-off, as the total computation time per run for $GA\_v2+GNN$ increased by almost a factor of four compared to $GA\_v2$.
 
 We continued the previous experiment by rerunning the $GA\_v2+GNN$ version and adding a new stop criterion: each run of the algorithm was terminated when its execution time reached the total execution time of the corresponding matching run associated to $GA\_v2$. We named this new hybrid version $GA\_v2+GNN*$. The comparison between $GA\_v2$ and $GA\_v2+GNN*$ enabled us to investigate whether the hybrid version could improve the values of the metrics for the same running times. 
 
 Table \ref{quality_tab2} contains the results of the comparison between  $GA\_v2$ and $GA\_v2+GNN*$. As before, we used the Welch's t-test and marked in bold the results that were statistically better, considering a $95\%$ confidence level.

 \begin{table}[!htb]
 	\caption{Comparative evaluation of the $GA\_v2$ and $GA\_v2+GNN*$ methods }
 	\label{quality_tab2}
 	\scalebox{0.77}{
 		\begin{tabular}{|l|ll|ll|ll|l|l|ll|}
 			\hline
 			\multirow{2}{*}{\begin{tabular}[c]{@{}l@{}}Metric\\ Method\end{tabular}} &
 			\multicolumn{2}{l|}{\begin{tabular}[c]{@{}l@{}}Fitness \\ value\end{tabular}} &
 			\multicolumn{2}{l|}{\begin{tabular}[c]{@{}l@{}}Soft \\ penalty\end{tabular}} &
 			\multicolumn{2}{l|}{\begin{tabular}[c]{@{}l@{}}Hard \\ penalty\end{tabular}} &
 			\begin{tabular}[c]{@{}l@{}}Number of\\ feasible \\ schedules\end{tabular} &
 			\begin{tabular}[c]{@{}l@{}}Number of\\ optimal\\ schedules\end{tabular} &
 			\multicolumn{2}{l|}{\begin{tabular}[c]{@{}l@{}}Crowding\\ distance\end{tabular}} \\ \cline{2-11} 
 			&
 			\multicolumn{1}{l|}{mean} &
 			max &
 			\multicolumn{1}{l|}{min} &
 			mean &
 			\multicolumn{1}{l|}{min} &
 			mean &
 			mean &
 			mean &
 			\multicolumn{1}{l|}{mean} &
 			max \\ \hline
 			\rotatebox{90}{$GA\_v2$} &
 			\multicolumn{1}{l|}{\begin{tabular}[c]{@{}l@{}}2920.5296\\ (0.1464)\end{tabular}} &
 			\begin{tabular}[c]{@{}l@{}}2920.7165\\ (0.0017)\end{tabular} &
 			\multicolumn{1}{l|}{\begin{tabular}[c]{@{}l@{}}19388.8333\\ (13.7633)\end{tabular}} &
 			\begin{tabular}[c]{@{}l@{}}19550.3685\\ (120.4645)\end{tabular} &
 			\multicolumn{1}{l|}{\begin{tabular}[c]{@{}l@{}}0.0\\ (0.0)\end{tabular}} &
 			\begin{tabular}[c]{@{}l@{}}0.1834\\ (0.1447)\end{tabular} &
 			\begin{tabular}[c]{@{}l@{}}173.8857\\ (16.2386)\end{tabular} &
 			\textbf{\begin{tabular}[c]{@{}l@{}}20.5428\\ (35.7008)\end{tabular}} &
 			\multicolumn{1}{l|}{\textbf{\begin{tabular}[c]{@{}l@{}}471.7008\\ (17.4030)\end{tabular}}} &
 			\textbf{\begin{tabular}[c]{@{}l@{}}542.8833\\ (22.9539)\end{tabular}} \\ \hline
 			\rotatebox{90}{$GA\_v2+GNN*$} &
 			\multicolumn{1}{l|}{\textbf{\begin{tabular}[c]{@{}l@{}}2920.6107\\ (0.0437)\end{tabular}}} &
 			\textbf{\begin{tabular}[c]{@{}l@{}}2920.7178\\ (0.0007)\end{tabular}} &
 			\multicolumn{1}{l|}{\textbf{\begin{tabular}[c]{@{}l@{}}19380.5\\ (8.2616)\end{tabular}}} &
 			\textbf{\begin{tabular}[c]{@{}l@{}}19421.6939\\ (16.4438)\end{tabular}} &
 			\multicolumn{1}{l|}{\begin{tabular}[c]{@{}l@{}}0.0\\ (0.0)\end{tabular}} &
 			\textbf{\begin{tabular}[c]{@{}l@{}}0.1051\\ (0.0434)\end{tabular}} &
 			\textbf{\begin{tabular}[c]{@{}l@{}}183.3333\\ (6.1952)\end{tabular}} &
 			\begin{tabular}[c]{@{}l@{}}4.8667\\ (8.0610)\end{tabular} &
 			\multicolumn{1}{l|}{\begin{tabular}[c]{@{}l@{}}463.7544\\ (5.1455)\end{tabular}} &
 			\begin{tabular}[c]{@{}l@{}}523.1833\\ (6.8035)\end{tabular} \\ \hline
 		\end{tabular}
 	}
 \end{table}
 
 The $GA\_v2+GNN*$ attained 
 superior results for the following metrics: average and maximum fitness value, minimum and mean soft penalty, mean hard penalty, and average number of feasible schedules, when compared to $GA\_v2$. On the other hand, $GA\_v2+GNN*$ obtained worse results for the average and maximum crowding distance and for the average number of optimal schedules.
 To conclude, the $GA\_v2+GNN*$ method generated less diverse timetables and fewer optimal solutions. However, in contrast to $GA\_v2$, the quality of the provided schedules was statistically higher, both when comparing all obtained timetables or only the best ones.

\section{Conclusions and future work} \label{conclusions_future_work}

The current paper investigates the impact of combining two fundamentally different optimization paradigms - a search heuristic based on populations in the form of a Genetic Algorithm and a data-driven Graph Neural Network - for solving a timetabling problem of great practical importance,  namely Staff Rostering. To the best of our knowledge, this study is the first to hybridize a Genetic Algorithm with a Graph Neural Network to address timetabling tasks. The GNN is capable of extracting and encapsulating general domain knowledge during training to enhance the quality of suboptimal schedules, while the GA explores the multi-modal solution space, exploiting the GNN accumulated knowledge in the form of a variation operator.
A thorough experimental analysis demonstrates that the hybrid method provides statistically significant improvements in both time efficiency and solution quality compared to the standalone methods. 

Regarding future directions, this study can be further extended to tackle other types of timetabling problems or more general scheduling tasks. Another promising direction focuses on devising a new technique to improve the generalization ability of the Graph Neural Network and investigating novel ways of hybridizing GNNs with meta-heuristics.

 \bibliographystyle{splncs04}
 \bibliography{bibliography}

@article{Asurveyofthenurserosteringsolutionmethodologies,
	title={A survey of the nurse rostering solution methodologies: The state-of-the-art and emerging trends},
	author={Ngoo, Chong Man and Goh, Say Leng and Sabar, Nasser R and Abdullah, Salwani and Kendall, Graham and others},
	journal={IEEE Access},
	volume={10},
	pages={56504--56524},
	year={2022},
	publisher={IEEE}
}

@article{Thecrowdingapproachtonichingingeneticalgorithms,
	title={The crowding approach to niching in genetic algorithms},
	author={Mengshoel, Ole J and Goldberg, David E},
	journal={Evolutionary computation},
	volume={16},
	number={3},
	pages={315--354},
	year={2008},
	publisher={MIT Press One Rogers Street, Cambridge, MA 02142-1209, USA journals-info~…}
}

@article{EverythingisconnectedGraphneuralnetworks,
	title={Everything is connected: Graph neural networks},
	author={Veli{\v{c}}kovi{\'c}, Petar},
	journal={Current Opinion in Structural Biology},
	volume={79},
	pages={102538},
	year={2023},
	publisher={Elsevier}
}

@article{Combinatorialoptimizationandreasoningwithgraphneuralnetworks,
	title={Combinatorial optimization and reasoning with graph neural networks},
	author={Cappart, Quentin and Ch{\'e}telat, Didier and Khalil, Elias B and Lodi, Andrea and Morris, Christopher and Veli{\v{c}}kovi{\'c}, Petar},
	journal={Journal of Machine Learning Research},
	volume={24},
	number={130},
	pages={1--61},
	year={2023}
}

@article{TowardsunderstandingconvergenceandgeneralizationofAdamW,
	title={Towards understanding convergence and generalization of AdamW},
	author={Zhou, Pan and Xie, Xingyu and Lin, Zhouchen and Yan, Shuicheng},
	journal={IEEE transactions on pattern analysis and machine intelligence},
	volume={46},
	number={9},
	pages={6486--6493},
	year={2024},
	publisher={IEEE}
}

@ARTICLE{scipy,
	author  = {Virtanen, Pauli and Gommers, Ralf and Oliphant, Travis E. and
	Haberland, Matt and Reddy, Tyler and Cournapeau, et al {SciPy 1.0 Contributors}},
	title   = {{{SciPy} 1.0: Fundamental Algorithms for Scientific
	Computing in Python}},
	journal = {Nature Methods},
	year    = {2020},
	volume  = {17},
	pages   = {261--272},
	adsurl  = {https://rdcu.be/b08Wh},
	doi     = {10.1038/s41592-019-0686-2},
}

@inproceedings{numba,
	title={Numba: A llvm-based python jit compiler},
	author={Lam, Siu Kwan and Pitrou, Antoine and Seibert, Stanley},
	booktitle={Proceedings of the Second Workshop on the LLVM Compiler Infrastructure in HPC},
	pages={1--6},
	year={2015}
}

@article{MaskedlabelpredictionUnifiedmessagepassingmodelforsemisupervisedclassification,
	title={Masked label prediction: Unified message passing model for semi-supervised classification},
	author={Shi, Yunsheng and Huang, Zhengjie and Feng, Shikun and Zhong, Hui and Wang, Wenjing and Sun, Yu},
	booktitle={Proceedings of the Thirtieth International Joint Conference on Artificial Intelligence},
	pages={1548--1554},
	year={2021},
	organization={International Joint Conferences on Artificial Intelligence Organization}
}

@inproceedings{Modelingrelationaldatawithgraphconvolutionalnetworks,
	title={Modeling relational data with graph convolutional networks},
	author={Schlichtkrull, Michael and Kipf, Thomas N and Bloem, Peter and Van Den Berg, Rianne and Titov, Ivan and Welling, Max},
	booktitle={European semantic web conference},
	pages={593--607},
	year={2018},
	organization={Springer}
}

@article{PyTorchGeometric,
	title={Fast graph representation learning with PyTorch Geometric},
	author={Fey, Matthias and Lenssen, Jan Eric},
	year={2019},
	url = "https://pytorch-geometric.readthedocs.io/en/latest/"
}

@inproceedings{oversmoothing,
	title={A comprehensive review of the oversmoothing in graph neural networks},
	author={Zhang, Xu and Xu, Yonghui and He, Wei and Guo, Wei and Cui, Lizhen},
	booktitle={CCF Conference on Computer Supported Cooperative Work and Social Computing},
	pages={451--465},
	year={2023},
	organization={Springer}
}

@misc{gurobi,
	author = {Gurobi Optimization, LLC},
	title = {Gurobi Optimizer Reference Manual},
	year = 2024,
	url = "https://www.gurobi.com"
}

@article{NofreelunchtheoremAreview,
	title={No free lunch theorem: A review},
	author={Adam, Stavros P and Alexandropoulos, Stamatios-Aggelos N and Pardalos, Panos M and Vrahatis, Michael N},
	journal={Approximation and optimization: Algorithms, complexity and applications},
	pages={57--82},
	year={2019},
	publisher={Springer}
}

@misc{Computationalresultsonnewstaffschedulingbenchmarkinstances,
	title={Computational results on new staff scheduling benchmark instances},
	author={Curtois, Tim and Qu, Rong},
	year={2014},
	publisher={NRP}
}

@article{Agnn-guidedpredict-and-searchframeworkformixed-integerlinearprogramming,
	title={A gnn-guided predict-and-search framework for mixed-integer linear programming},
	author={Han, Qingyu and Yang, Linxin and Chen, Qian and Zhou, Xiang and Zhang, Dong and Wang, Akang and Sun, Ruoyu and Luo, Xiaodong},
    booktitle={ICLR},
    year={2023}
}

@article{RL-MILPSolver:areinforcementlearningapproachforsolvingmixed-integerlinearprogramswithgraphneuralnetworks,
	title={RL-MILP Solver: a reinforcement learning approach for solving mixed-integer linear programs with graph neural networks},
	author={Lee, Tae-Hoon and Kim, Min-Soo},
	journal={arXiv preprint arXiv:2411.19517},
	year={2024}
}

@article{Combinatorialoptimizationwithautomatedgraphneuralnetworks,
	title={Combinatorial optimization with automated graph neural networks},
	author={Liu, Yang and Zhang, Peng and Gao, Yang and Zhou, Chuan and Li, Zhao and Chen, Hongyang},
	journal={arXiv preprint arXiv:2406.02872},
	year={2024}
}

@inproceedings{Neurallargeneighborhoodsearch,
	title={Neural large neighborhood search},
	author={Nair, Vinod and Alizadeh, Mohammad and others},
	booktitle={Learning Meets Combinatorial Algorithms at NeurIPS2020},
	year={2020}
}

@inproceedings{Alearninglargeneighborhoodsearchforthestaffrerosteringproblem,
	title={A learning large neighborhood search for the staff rerostering problem},
	author={Oberweger, Fabio F and Raidl, G{\"u}nther R and R{\"o}nnberg, Elina and Huber, Marc},
	booktitle={International Conference on Integration of Constraint Programming, Artificial Intelligence, and Operations Research},
	pages={300--317},
	year={2022},
	organization={Springer}
}

@inproceedings{AutomatedPersonnelSchedulingwithReinforcementLearningandGraphNeuralNetworks,
	title={Automated Personnel Scheduling with Reinforcement Learning and Graph Neural Networks},
	author={Platten, Benjamin and Macfarlane, Matthew and Graus, David and Mesbah, Sepideh},
	booktitle={HR@ RecSys},
	year={2022}
}

@inproceedings{FasterLargerStrongerOptimallySolvingEmployeeSchedulingProblemswithGraphNeuralNetworks,
	title={Faster, Larger, Stronger: Optimally Solving Employee Scheduling Problems with Graph Neural Networks},
	author={Nguyen, Duc Huy and Truong, Tran Quoc An and Tran-Thanh, Long},
	booktitle={International Symposium on Information and Communication Technology},
	pages={141--151},
	year={2024},
	organization={Springer}
}

@article{Ageneticalgorithmforthepersonneltaskreschedulingproblemwithtimepreemption,
	title={A genetic algorithm for the personnel task rescheduling problem with time preemption},
	author={Borgonjon, Tessa and Maenhout, Broos},
	journal={Expert Systems with Applications},
	volume={238},
	pages={121868},
	year={2024},
	publisher={Elsevier}
}

@article{Apreventivereactiveapproach,
	title={A preventive-reactive approach for nurse scheduling considering absenteeism and nurses’ preferences},
	author={Otero-Caicedo, Ricardo and Casas, Carlos Eduardo Montoya and Jaimes, Carolina Barajas and Garz{\'o}n, Cristian Felipe Guzm{\'a}n and Vergel, Edwin Andr{\'e}s Y{\'a}{\~n}ez and Vald{\'e}s, Juli{\'a}n Camilo Zabala},
	journal={Operations Research for Health Care},
	volume={38},
	pages={100389},
	year={2023},
	publisher={Elsevier}
}

@article{Nurserosteringwithfatiguemodelling,
	title={Nurse rostering with fatigue modelling: Incorporating a validated sleep model with biological variations in nurse rostering},
	author={Klyve, Kjartan Kastet and Senthooran, Ilankaikone and Wallace, Mark},
	journal={Health care management science},
	volume={26},
	number={1},
	pages={21--45},
	year={2023},
	publisher={Springer}
}

@article{Ahybridintegerprogrammingandartificialbeecolonyalgorithmforstaffschedulingincallcenters,
	title={A hybrid integer programming and artificial bee colony algorithm for staff scheduling in call centers},
	author={Xu, Yue and Wang, Xiuli},
	journal={Computers \& Industrial Engineering},
	volume={171},
	pages={108312},
	year={2022},
	publisher={Elsevier}
}

@article{Astochasticintegerprogrammingapproachtoreservestaffschedulingwithpreferences,
	title={A stochastic integer programming approach to reserve staff scheduling with preferences},
	author={Perreault-Lafleur, Carl and Carvalho, Margarida and Desaulniers, Guy},
	journal={International Transactions in Operational Research},
	volume={32},
	number={1},
	pages={289--313},
	year={2025},
	publisher={Wiley Online Library}
}

@article{MaximizingShiftPreferenceforNurseRosteringScheduleUsingIntegerLinearProgrammingandGeneticAlgorithm,
	title={Maximizing Shift Preference for Nurse Rostering Schedule Using Integer Linear Programming and Genetic Algorithm},
	author={Mohd Razali, Siti Noor Asyikin Binti and Tamilarasan, Thesigan Achari A and Basri, Batrisyia Binti and Bin Arbin, Norazman and others},
	journal={International Journal of Advanced Computer Science \& Applications},
	volume={16},
	number={5},
	year={2025}
}

@article{Amatheuristicbasedsolutionapproachforanextendednurserosteringproblemwithskillsandunits,
	title={A mat-heuristic based solution approach for an extended nurse rostering problem with skills and units},
	author={Turhan, Aykut Melih and Bilgen, Bilge},
	journal={Socio-Economic Planning Sciences},
	volume={82},
	pages={101300},
	year={2022},
	publisher={Elsevier}
}

@inproceedings{Amultiobjectiveevolutionaryapproachtoprofessionalcoursetimetabling,
	title={A multi-objective evolutionary approach to professional course timetabling: A real-world case study},
	author={Hafsa, Mounir and Wattebled, Pamela and Jacques, Julie and Jourdan, Laetitia},
	booktitle={2021 IEEE Congress on Evolutionary Computation (CEC)},
	pages={997--1004},
	year={2021},
	organization={IEEE}
}

@article{SolvingamultiobjectiveprofessionaltimetablingproblemusingevolutionaryalgorithmsatMandarineAcademy,
	title={Solving a multiobjective professional timetabling problem using evolutionary algorithms at Mandarine Academy},
	author={Hafsa, Mounir and Wattebled, Pamela and Jacques, Julie and Jourdan, Laetitia},
	journal={International Transactions in Operational Research},
	volume={32},
	number={1},
	pages={244--269},
	year={2025},
	publisher={Wiley Online Library}
}

@article{Multiobjectivehybridoptimizationsfordesigningcourseschedulesbasedonoperatingcostsandresourceutilization,
	title={Multi-objective hybrid optimizations for designing course schedules based on operating costs and resource utilization},
	author={Thepphakorn, Thatchai and Pongcharoen, Pupong and Vitayasak, Srisatja},
	journal={Annals of Operations Research},
	pages={1--68},
	year={2024},
	publisher={Springer}
}

@article{Noveloperatorsforquantumevolutionaryalgorithminsolvingtimetablingproblem,
	title={Novel operators for quantum evolutionary algorithm in solving timetabling problem},
	author={Tayarani-N, Mohammad-H},
	journal={Evolutionary Intelligence},
	volume={14},
	number={4},
	pages={1869--1893},
	year={2021},
	publisher={Springer}
}

@inproceedings{AcceleratingModelSolvingforIntegratedOptimizationofTimetablingandVehicleSchedulingbasedonGraphConvolutionalNetwork,
	title={Accelerating Model Solving for Integrated Optimization of Timetabling and Vehicle Scheduling based on Graph Convolutional Network},
	author={Liu, YeCheng and Chen, Xu and Xu, YongXin and Xiang, Dong and Mo, LinJian},
	booktitle={2023 IEEE 26th International Conference on Intelligent Transportation Systems (ITSC)},
	pages={880--886},
	year={2023},
	organization={IEEE}
}

@article{Reinforcementlearningforscalabletraintimetablereschedulingwithgraphrepresentation,
	title={Reinforcement learning for scalable train timetable rescheduling with graph representation},
	author={Yue, Peng and Jin, Yaochu and Dai, Xuewu and Feng, Zhenhua and Cui, Dongliang},
	journal={IEEE Transactions on Intelligent Transportation Systems},
	volume={25},
	number={7},
	pages={6472--6485},
	year={2024},
	publisher={IEEE}
}

@article{Solvingtherailwaytimetablereschedulingproblemwithgraphneuralnetworks,
	title={Solving the railway timetable rescheduling problem with graph neural networks},
	author={Huang, Ping and Peng, Zihuan and Li, Zhongcan and Peng, Qiyuan},
	journal={Railway Engineering Science},
	pages={1--22},
	year={2025},
	publisher={Springer}
}

@article{Barriersfortheperformanceofgraphneuralnetworksindiscreterandomstructures,
	title={Barriers for the performance of graph neural networks (GNN) in discrete random structures},
	author={Gamarnik, David},
	journal={Proceedings of the National Academy of Sciences},
	volume={120},
	number={46},
	pages={e2314092120},
	year={2023},
	publisher={National Academy of Sciences}
}

@incollection{Geneticalgorithmsandadaptation,
  title={Genetic algorithms and adaptation},
  author={Holland, John H},
  booktitle={Adaptive control of ill-defined systems},
  pages={317--333},
  year={1984},
  publisher={Springer}
}

\end{document}